\documentclass{article}
\PassOptionsToPackage{numbers,sort&compress}{natbib}
\usepackage[final]{tackling_climate_workshop_style}
\usepackage[utf8]{inputenc}
\usepackage[T1]{fontenc}
\usepackage{courier}
\usepackage{amsmath,amssymb}
\usepackage{graphicx,booktabs,array,tabularx,makecell}
\usepackage{microtype}
\usepackage{placeins}
\usepackage{float}
\usepackage{etoolbox}
\makeatletter
\patchcmd{\@makecaption}{#1: #2}{#1. #2}{}{\PackageError{paper}{Caption punctuation patch failed}{}}
\patchcmd{\@makecaption}{#1: #2}{#1. #2}{}{\PackageError{paper}{Caption punctuation patch failed}{}}
\renewcommand{\@noticestring}{Tackling Climate Change with Machine Learning workshop at NeurIPS 2026.}
\makeatother
\usepackage{xurl}
\usepackage{hyperref}
\hypersetup{colorlinks=true,linkcolor=black,citecolor=black,urlcolor=black,pdftitle={Sensor Diagnosis for Fuel and Emissions Monitoring in Heavy-Duty Diesel Vehicles},pdfauthor={Tejas Dahiya and Saurabh Kumar Gupta}}
\title{Sensor Diagnosis for Fuel and Emissions\protect\\Monitoring in Heavy-Duty Diesel Vehicles}
\author{Tejas Dahiya\\University of Chicago\And Saurabh Kumar Gupta\\University of Wisconsin--Madison}
\newcommand{\cotwo}{CO\textsubscript{2}}
\newcommand{\pct}{\%}
\newcolumntype{L}[1]{>{\raggedright\arraybackslash}p{#1}}
\newcolumntype{C}[1]{>{\centering\arraybackslash}p{#1}}
\begin{document}
\maketitle
\begin{abstract}
Faulty sensor readings can distort estimates of fuel use and carbon dioxide emissions from diesel engines. We evaluate two procedures that use relationships learned from healthy measurements to diagnose faults and guide fuel correction. The diagnosis procedure tests whether replacing a suspected faulty reading with an estimated healthy value helps a model predict another sensor's recorded reading more accurately. It then predicts the fault type and decides whether to report the diagnosis. We evaluate five synthetic fault types on 217 steady-state operating points from a heavy-duty diesel engine, excluding all measurements at selected engine speeds from training. Across 31,500 fault-injected groups of 20 operating points, it issues diagnoses for 77.3\pct\ of groups, with 92.1\pct\ of diagnoses issued on these groups correct. The fuel correction procedure uses a different sensor-selection rule at individual operating points, testing how each replacement changes the largest prediction error across all six sensors. Fuel correction requires a fault alarm and engine speed and torque close to the training conditions. A fuel reading is replaced only when its replacement reduces the largest prediction error sufficiently more than replacements for the other sensors. Assuming faults in 1\pct\ of one-hour scenarios, equally likely sensors and fault types, and equal time at each measured condition, selected fuel corrections reduce mean absolute error in calculated combustion \cotwo\ by 83.8\pct. On this engine map, learned relationships support complete sensor diagnoses and fuel corrections that improve emissions accounting while preserving unaffected readings.
\end{abstract}

\section{Introduction}
Heavy-duty vehicles produce about one quarter of road-transport \cotwo\ emissions in the European Union \citep{eea}. The International Energy Agency's Current Policies Scenario projects that only about 10\pct\ of the global truck fleet will be electric by 2035 \citep{iea}, so diesel vehicles in service will continue to require fuel and emissions monitoring. Such monitoring relies on sensor readings from the intake, combustion, and exhaust systems. Errors in these readings can affect engine control and calculations of fuel use and emissions.

Relationships learned from healthy measurements provide one way to identify faulty sensors. A faulty reading can make the model's predictions for healthy sensors wrong, so the sensor whose reading differs most from its predicted value may be healthy. Earlier work addresses this problem by reconstructing measurements and testing whether replacing a suspected reading restores agreement with a healthy process model \citep{frank,dunia,qin,alcala}. Automotive studies use neural networks and decision rules for detection, identification, and compensation \citep{capriglione}, or principal-component and autoencoder reconstruction \citep{schmitz2024scalable}. Zhang and Li predict nitrogen-oxide readings to reconstruct a faulty signal during transient diesel-engine operation \citep{zhang}.

We evaluate how sensor identification affects the accuracy and frequency of complete diagnoses and whether selected fuel corrections improve emissions estimates. Diagnosis combines sensor identification, fault-type prediction, and a model deciding whether to issue the result, following work on withholding uncertain predictions \citep{chow,selective}. We compare identification methods at engine speeds absent from training. A distinct procedure selects fuel corrections at individual operating points, without the fault-type or reporting models, and measures their effect on calculated \cotwo.

\section{Method}
\label{sec:method}
\begin{figure}[htbp]
\centering
\includegraphics[width=\linewidth]{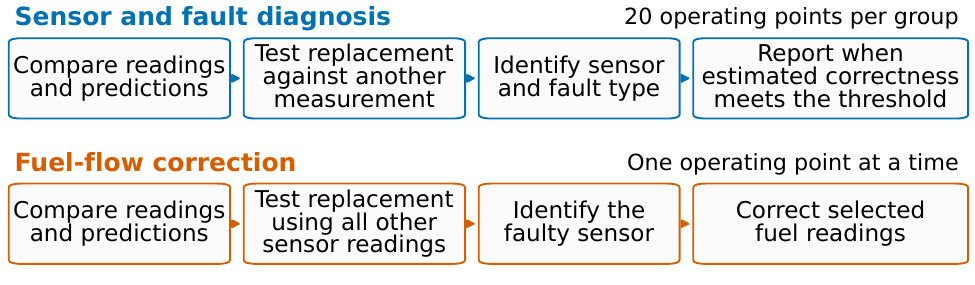}
\caption{Diagnosis combines 20 operating points, while fuel correction evaluates each point.}
\label{fig:methods}
\end{figure}

Each operating point contains sensor readings measured at one engine speed and torque. The diagnosis procedure combines 20 distinct points (Figure~\ref{fig:methods}) and monitors intake-manifold temperature, intake-manifold pressure, exhaust-gas-recirculation rate, intake-air mass flow, and engine-out nitrogen-oxide concentration. Speed and torque describe the operating condition, and neither receives synthetic faults. For each monitored sensor, second-order polynomial ridge regression predicts its healthy reading from speed, torque, and the other sensors \citep{ridge}.

To identify unusual disagreement, we scale each prediction error by its standard deviation on healthy data and the variation among predictions from an ensemble of regression trees \citep{extratrees}. A point qualifies for the alarm calculation when its largest adjusted sensor error exceeds a healthy-training threshold and its speed and torque are close to training conditions. The group raises an alarm when its count of qualifying points reaches a threshold selected using healthy groups constructed from training observations (Appendix~\ref{app:alarm}).

Sensor identification tests whether a suspected reading contributes to another sensor's prediction error. We first estimate a replacement from speed, torque, and one other sensor. A second model checks its effect by predicting a different sensor from speed, torque, the suspected reading, and a fourth sensor. All four sensor roles are distinct, so the measurement used to estimate the replacement is excluded from the checking model's inputs and target.

We compare the checking model's error before and after inserting the replacement, scaling the reduction by the variation in its healthy prediction errors. Summing these reductions across qualifying points gives a total for each suspected sensor. We select the sensor with the largest total. Appendix~\ref{app:methods-source} specifies how the sensor combinations and models are chosen.

A random forest predicts one of five fault types from summaries of the selected sensor's readings and prediction errors, together with measures of sensor-selection strength and consistency across the group \citep{forest}. These errors come from a separate polynomial ridge model using speed and torque alone.

For training and evaluation, a diagnosis is correct only if both the sensor and fault type are identified correctly, the injection changes a reading, the group raises an alarm, and a point passes the speed and torque check. Logistic regression estimates this correctness from the alarm, operating-condition check, sensor-selection summaries, and fault-type predictions. Diagnoses on healthy groups or unchanged injections are labelled incorrect during training. Reporting depends on the estimated correctness reaching a selected threshold. The alarm informs this estimate but is not an additional requirement for reporting.

Within each split's training observations, four partitions supply the groups used to fit the fault-type and reporting models. All groups from a partition stay together, and models scoring them exclude that partition. An eligible reporting threshold must issue at least 20 training diagnoses and have a lower 95\pct\ Wilson confidence bound of at least 90\pct\ for their correctness. We select the eligible threshold issuing the most diagnoses, then fit the final models and keep the thresholds fixed during testing. Appendix~\ref{app:methods-reporting} gives the nested fitting sequence.

\section{Experiments and results}
\subsection{Data and evaluation}
We use 217 healthy steady-state operating points from a 12-litre diesel engine. Each of seven splits reserves three of the 21 speeds for testing, leaving 186 training and 31 test observations. Runs use seeds 42, 123, and 2026, with all fitting and threshold selection confined to training observations.

Synthetic faults add offsets, change the measurement scale, add noise, replace intermittent readings with zero, or cap readings above an upper limit. Development and testing use different magnitudes. Each case applies one fault type to one sensor in the development or test observations before sampling 20 distinct points without replacement. The 31,500 groups can reuse the 217 measured points.

The system evaluates test measurements without access to their healthy reference values or injected-fault records. These records supply training labels and allow us to check the test predictions. We report alarm and diagnosis rates and correctness on fault-injected groups, as defined in Section~\ref{sec:method}. Diagnoses on healthy groups are assessed separately, as are results restricted to injections that change a reading (Table~\ref{tab:diagnosis}).

\subsection{Sensor diagnosis}
Across the 31,500 fault-injected groups, the system raises alarms for 80.0\pct\ and issues diagnoses for 77.3\pct. Among diagnoses issued on these groups, 92.1\pct\ are correct. Sensor identification among groups raising alarms is correct in 96.7\pct\ of cases. The generated examples give unequal weight to the five fault types. Giving each type equal weight lowers reporting to 70.4\pct\ and correctness to 90.5\pct\ (Table~\ref{tab:diagnosis}).

\begin{table}[htbp]
\caption{Diagnosis on fault-injected groups at engine speeds excluded from training. Correctness is measured among diagnoses issued on those groups; healthy groups are evaluated separately.}
\label{tab:diagnosis}
\centering
\begin{tabular}{lrrr}
\toprule
Evaluated groups & \makecell{Alarm\\raised} & \makecell{Diagnosis\\issued} & \makecell{Diagnosis\\correct}\\
\midrule
All generated injections & 80.0\pct & 77.3\pct & 92.1\pct\\
Equally likely fault types & 72.6\pct & 70.4\pct & 90.5\pct\\
Injections that change a reading & 82.3\pct & 79.6\pct & 92.3\pct\\
\bottomrule
\end{tabular}
\end{table}

\begin{figure}[htbp]
\centering
\includegraphics[width=\linewidth]{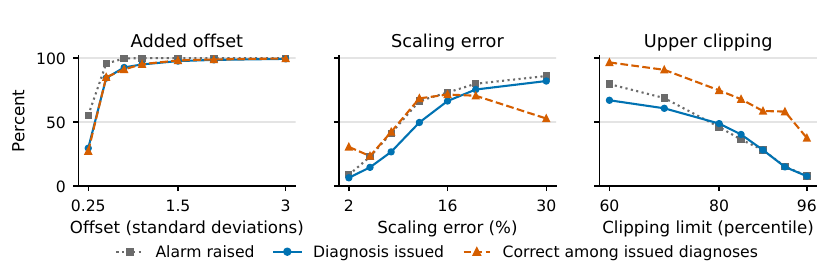}
\caption{Alarm frequency, diagnosis frequency, and correctness among issued diagnoses as fault magnitude changes, pooling five sensors, three seeds, and seven test splits. Offsets are expressed in training standard deviations and clipping limits in training percentiles. Models and thresholds remain fixed, so the curves show their responses to changing fault magnitude.}
\label{fig:severity}
\end{figure}

This overall result varies substantially by fault type. Correctness among issued diagnoses reaches 96.4\pct\ for offsets, 99.98\pct\ for noise, and 96.0\pct\ for intermittent zero readings, but falls to 63.1\pct\ for scaling errors and 65.2\pct\ for capped readings. Figure~\ref{fig:severity} examines these differences as fault magnitude changes.

We also test healthy groups to measure how often the system diagnoses a fault when none was injected. Of 2,520 healthy groups, 171 raise alarms and 169 receive diagnoses, giving rates of 6.8\pct\ and 6.7\pct. All diagnoses issued on healthy groups occur in one test split for seeds 42 and 123, with none in the other 19 seed-and-split combinations.

\subsection{Comparison of sensor-identification methods}
We compare identification methods on the same test groups and alarm procedure, retraining the fault-type and reporting models for each one. Replacement testing and the largest adjusted prediction error produce similar reporting and correctness rates (Table~\ref{tab:comparison}). Accounting for correlated errors reduces diagnoses on healthy groups to 0.2\pct, while issuing fewer fault diagnoses with lower correctness. The comparison therefore shows different tradeoffs between reporting, correctness, and healthy-group errors. Appendix~\ref{app:comparison} specifies the implementations and detailed results.

\begin{table}[htbp]
\caption{Identification methods with the same alarms and newly fitted fault-type and reporting models. Reporting and correctness concern fault-injected groups; the final column gives diagnoses on healthy groups.}
\label{tab:comparison}
\centering
\begin{tabular}{lrrr}
\toprule
Sensor-identification method & \makecell{Diagnosis\\issued} & \makecell{Diagnosis\\correct} & \makecell{Healthy groups\\diagnosed}\\
\midrule
Test estimated replacements & 77.3\pct & 92.1\pct & 6.7\pct\\
Largest adjusted prediction error & 78.0\pct & 91.8\pct & 6.8\pct\\
Account for correlated errors & 63.4\pct & 87.1\pct & 0.2\pct\\
Principal-component reconstruction & 74.6\pct & 88.8\pct & 11.5\pct\\
Autoencoder reconstruction & 63.1\pct & 91.8\pct & 3.0\pct\\
\bottomrule
\end{tabular}
\end{table}

\FloatBarrier
\subsection{Fuel correction and calculated emissions}
We evaluate fuel correction at individual operating points, adding separately measured fuel flow to the five engine measurements and testing faults in all six sensors. Each sensor's healthy value is estimated from speed, torque, and the other five readings, excluding brake-specific fuel consumption and injected fuel quantity because these quantities directly describe fuel use. We insert each estimate and recalculate the largest adjusted prediction error across the sensors. The sensor whose replacement produces the greatest proportional reduction is selected as faulty.

Fuel flow is corrected only when an alarm is raised, speed and torque pass the operating-condition check, and fuel flow is selected as faulty. Its proportional error reduction must also exceed the next-largest reduction by a margin selected using development data. These checks determine correction directly, without the diagnosis experiment's fault-type classifier or reporting model. The fuel experiment uses the same engine-speed splits and three seeds, with three realizations of stochastic faults.

For emissions accounting, we distribute an assumed hour across the 217 operating points and convert fuel use to combustion \cotwo\ using 3.1863~kg \cotwo\ per kg diesel \citep{ipcc}. We measure the absolute difference from the total calculated with healthy reference readings and average it across assumed healthy and faulty hours. With 99\pct\ healthy hours, 1\pct\ faulty hours, equally likely sensors and fault types, and equal time at each point, correcting selected readings reduces error from 0.0529 to 0.00855~kg \cotwo\ per engine-hour, or 83.8\pct.

The comparison treatments show how the choice of readings affects this result. Replacing every fuel reading gives an error of 0.3488~kg, whereas correcting only readings changed by the injected faults, using the same estimated healthy values, gives 0.000325~kg. Weighting operating time toward low or high loads gives reductions of 83.2\pct\ and 83.8\pct, respectively.

\begin{figure}[H]
\centering
\includegraphics[width=\linewidth]{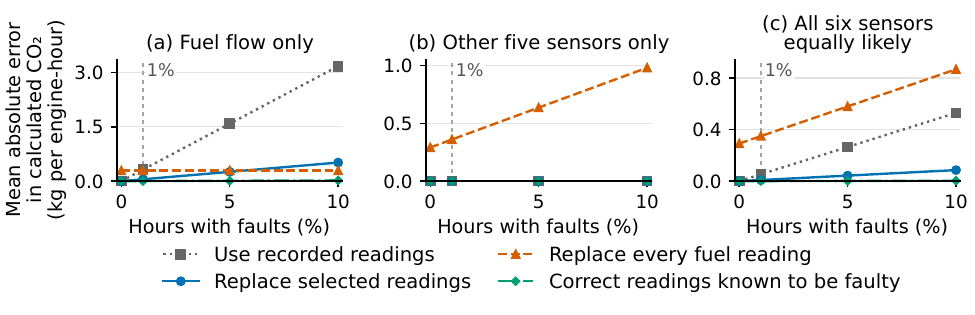}
\caption{Mean absolute error in calculated combustion \cotwo\ as the assumed fraction of faulty one-hour scenarios increases. The vertical markers identify the 1\pct\ assumption used for the headline result. Panels consider fuel flow, the other five sensors, or all six sensors with equal fault probability. Curves average three seeds with equal operating time and fixed correction rules. Correcting readings known to be faulty uses the same estimated replacements. Panel-specific vertical ranges show the different error levels; Appendix~\ref{app:climate} specifies the weights.}
\label{fig:climate-frequency}
\end{figure}

The procedure corrects 7,803 of the 11,591 fuel readings changed by injected faults. None of these corrections increases absolute fuel-flow error, and all 82,804 unaffected fuel readings are left unchanged.

\FloatBarrier
\section{Discussion and limitations}
On this engine map, replacement testing and the largest adjusted prediction error produce similar reporting and correctness rates. The comparison assesses identification rules within one diagnosis procedure, without reproducing complete published systems. Convergence warnings in 19 of 21 autoencoder fits further restrict that comparison.

In the fuel experiment, replacing every reading adds estimation error to unaffected measurements. At the assumed fault frequency, this error outweighs the benefit of correcting faulty readings, explaining why selective correction improves accounting accuracy.

Both diesel experiments use synthetic faults on one steady-state engine map. Groups and seeds reuse measurements, and component studies inform model choices (Appendix~\ref{app:components}). Some injections produce negative fuel readings that remain when uncorrected. Other diesel engines, transient operation, long-term degradation, and observed failures remain untested. The separately fitted turbofan experiment in Appendix~\ref{app:turbofan} evaluates detection and reconstruction under a different procedure. Physical engine changes or model errors can cause alarms, which therefore do not confirm sensor failures.

The errors on healthy groups concentrate at particular speeds, while scaling errors and capped readings remain harder to diagnose. For load-dependent offsets, stuck readings, and simultaneous faults in two sensors, the system reports one of its five training fault types for 90.2\pct\ to 98.5\pct\ of groups, so withholding some diagnoses does not reliably recognize these unfamiliar faults.

The accounting results depend on assumed fault frequencies and operating durations. Physical fuel savings and emissions reductions were not measured. Evaluation in vehicle operation requires observed sensor failures and measured operating histories.

\section{Conclusion}
On the tested engine map, relationships learned from healthy measurements support sensor diagnosis and improve fuel-based \cotwo\ accounting while preserving healthy fuel readings. Extending these results to diesel vehicles in service could improve the accuracy of reported fuel use and emissions. Future work should evaluate both procedures with observed sensor failures across different engines and changing speeds and loads, to establish whether diagnoses remain accurate and fuel correction improves \cotwo\ estimates during vehicle operation.

\clearpage
\setlength{\bibsep}{4pt plus 1pt minus 0.5pt}
\bibliographystyle{tccml_nocolons}
\bibliography{references}
\clearpage
\appendix
\section{Engine measurements and synthetic faults}
\label{app:data_faults}

The measurements come from a 12-litre, six-cylinder heavy-duty diesel engine tested on a dynamometer. The dataset contains 217 steady-state operating points at 21 nominal engine speeds. Recorded speed ranges from 549.9996 to 2150.001 revolutions per minute and torque from $-0.042$ to 2606.458 newton-metres. The small negative torque minimum is effectively zero at this scale. Each row records a distinct combination of speed and torque, together with the corresponding sensor measurements. The rows describe controlled engine conditions and contain no information about elapsed time between observations.

No synthetic faults are added to engine speed or torque in the primary diagnosis experiment. The five measurements examined for faults are intake-manifold temperature, intake-manifold pressure, exhaust-gas-recirculation rate, intake-air mass flow, and engine-out nitrogen oxide concentration. Their recorded ranges are respectively 31.0--69.9 degrees Celsius, 1.03--3.84 bar, 3.29--43.38 percent, 146.19--1632.78 kilograms per hour, and 105.85--1579.94 parts per million. These quantities describe related intake and combustion processes, allowing disagreement among measurements to provide evidence about sensor errors. Their original recorded values are treated as healthy. The experimental fault labels are obtained from the changes applied to those values. These injected reading errors differ from physical engine malfunctions studied in other experiments. In a Euro~VI truck test, simulated intake-system malfunctions increased fuel consumption by 16--43\pct\ \citep{huang}.

Table~\ref{tab:data_columns} maps the measurement names to the columns in the engine map used for these experiments. The separately measured fuel flow is included for the accounting experiment and is described in Appendix~\ref{app:climate}.

\begin{table}[htbp]
\centering
\caption{Measurement columns in the engine map used for these experiments. Speed and torque describe the operating condition, the next five measurements are monitored in the primary diagnosis, and fuel flow is added for accounting.}
\label{tab:data_columns}
\begin{tabular}{@{}p{0.19\linewidth}p{0.49\linewidth}p{0.24\linewidth}@{}}
\toprule
Data column & Measurement & Unit \\
\midrule
\texttt{Engine Speed} & Engine speed & Revolutions per minute \\
\texttt{Torque} & Engine torque & Newton-metres \\
\texttt{T\_IM} & Intake-manifold temperature & Degrees Celsius \\
\texttt{P\_IM} & Intake-manifold pressure & Bar \\
\texttt{EGR\_Rate} & Exhaust-gas-recirculation rate & Percent \\
\texttt{MF\_IA} & Intake-air mass flow & Kilograms per hour \\
\texttt{NOx\_EO} & Engine-out nitrogen oxide concentration & Parts per million \\
\texttt{MF\_FUEL} & Fuel mass flow & Kilograms per hour \\
\bottomrule
\end{tabular}
\end{table}
\FloatBarrier

Let $y_{ij}^{\mathrm{healthy}}$ be the healthy reference value of sensor $j$ at operating point $i$, and let $s_j$ be its sample standard deviation among the observations used for the model fit. An offset changes the value to $y_{ij}^{\mathrm{healthy}}+d\lambda s_j$, where $\lambda$ sets its size and $d\in\{-1,1\}$ its direction. A scaling error changes it to $y_{ij}^{\mathrm{healthy}}(1+dg)$, where $g$ is the fractional change. Added noise changes it to $y_{ij}^{\mathrm{healthy}}+\epsilon_i$, with independently drawn $\epsilon_i\sim\mathcal{N}(0,(ns_j)^2)$ and multiplier $n$. Clipping changes it to $\min(y_{ij}^{\mathrm{healthy}},c_{j,p})$, where $c_{j,p}$ is the $p$th percentile of healthy training values. An intermittent zero reading replaces the value with zero with probability $r$ and otherwise leaves it unchanged. Each fault acts on one sensor at a time. The resulting readings, denoted $y_{ij}$ below, are the inputs evaluated by the diagnostic models. Speed, torque, and the other sensor readings remain available.

Table~\ref{tab:fault_grid} gives the strengths used to train the fault classifier and reporting model and those reserved for testing. The sample standard deviations and clipping limits are recalculated from the fitting observations in every applicable split. Both signs are tested for offsets and scaling errors. Noise and intermittent zeros use two independent realizations per sensor and strength. Each realization is determined by a reproducible seed derived from the experiment seed, split, sensor, fault type, strength, direction, and realization identifier.

\begin{table}[H]
\centering
\caption{Fault strengths used in development and testing. Offset and noise sizes are multiples of the sensor's training standard deviation. Clipping limits are training percentiles.}
\label{tab:fault_grid}
\begin{tabular}{lll}
\toprule
Fault type & Development & Test \\
\midrule
Offset & $0.5,\ 1.0,\ 2.0$ & $0.75,\ 1.5,\ 3.0$ \\
Scaling error & $5,\ 12,\ 20\%$ & $8,\ 16\%$ \\
Added noise & $1.0,\ 2.0$ & $1.5,\ 3.0$ \\
Clipping & 80th, 88th & 84th, 92nd \\
Intermittent zeros & $15,\ 30\%$ & $10,\ 40\%$ \\
\bottomrule
\end{tabular}
\end{table}
\FloatBarrier

Clipping and intermittent zeros can leave some or all sampled readings unchanged. The implementation records the operating points at which the resulting value differs from the healthy value by more than $10^{-12}$ in absolute units. A group has an observable fault when at least one of its readings changes. The main evaluation includes every generated group. A separate evaluation restricts the denominator to groups with an observable change. During training, groups with no changed reading are labelled incorrect for reporting-model training and are excluded from fitting the fault-type classifier. Healthy groups receive the same label.

\section{Data splits and measurement groups}
\label{app:splits}

\begin{figure}[htbp]
\centering
\includegraphics[width=\linewidth]{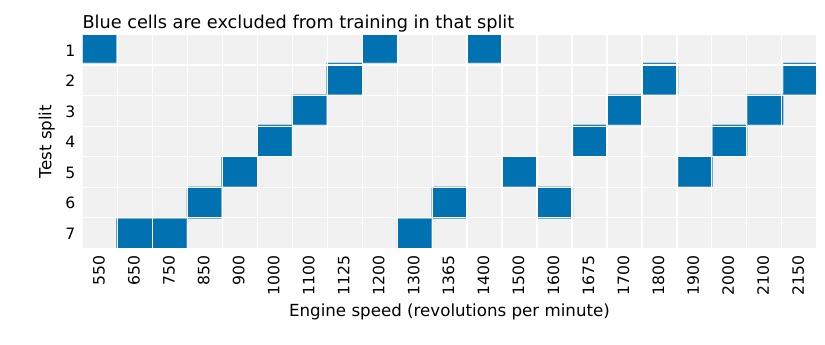}
\caption{The dark cells mark speeds at which all measurements are reserved for testing in each split. Each of the 21 speeds appears in one test split, and all three seeds use these assignments.}
\label{fig:splits}
\end{figure}

Recorded engine speeds are rounded to the nearest whole revolution per minute to obtain the 21 nominal speed levels. Seven training and test splits use the grouped cross-validation implementation in scikit-learn without shuffling, keeping each speed level intact and balancing test-portion sizes. Figure~\ref{fig:splits} shows their assignments. Every split uses 186 observations for training and 31 for testing, with all measurements at three engine speeds reserved for testing. Each nominal speed appears in a test portion once. All fitted models, scales, clipping limits, and selected thresholds for that split use its training observations.

Within each outer training split, the same grouped splitting rule divides its measured observations into four development partitions, keeping all observations at a nominal speed together. These are subdivisions of the training observations, rather than a separate measured dataset. To evaluate one partition, we fit the healthy-prediction and sensor-identification models on the other three. We then add synthetic faults to the reserved observations using the development strengths in Table~\ref{tab:fault_grid} and evaluate them with those fitted models. Repeating this step across the four partitions supplies the groups used to train the fault-type and reporting models. The procedure repeats with seeds 42, 123, and 2026. Speed assignments remain fixed, while seeds vary synthetic faults, sampled groups, and randomized model fitting.

The primary diagnosis combines 20 distinct operating points. A fault is first applied to the complete development or test portion and the system calculates the evidence at each operating point. Groups are then sampled from those already-scored rows. This preserves the same corrupted value and scores at each point whenever a row appears in several groups under the same fault realization. Each group draws its rows uniformly without replacement from one portion. Fifteen groups are drawn per sensor, strength, direction, and realization, and 120 groups are drawn from the unchanged measurements. The number of groups differs across fault types because the tested strengths, directions, and realizations differ. The overall results reflect these unequal numbers of test groups. We also report results for each fault type and recalculate the overall results with all five fault types weighted equally.

The group-size experiment uses 5, 10, and 20 rows. It compares the same uniform sampling with groups built around a randomly selected row. For the latter, the selected row and its nearest rows in standardized speed and torque are included. The speed and torque standardization uses training means and standard deviations. Groups never mix observations from opposite sides of a training or test split. Different groups can nevertheless share physical rows. Their number measures the number of generated diagnostic cases, while the number of independently measured operating points remains 217.

All groups derived from one development partition remain together during the further classifier fits used to select the reporting threshold. Section~\ref{app:reporting} specifies these fits and the treatment of healthy groups and injections with no changed reading. Appendix~\ref{app:earlier} describes the component comparisons used to select the model design.

\section{Healthy predictions and fault alarms}
\label{app:alarm}

For each sensor, a regression model predicts its healthy value from speed, torque, and the other four monitored measurements. Each input is standardized using the fitting data. The transformed inputs include their first-order terms, squares, and pairwise products. Ridge regression with penalty parameter 1.0 then fits the relationship~\citep{ridge}. The sensor being predicted is excluded from its input vector. Let $h_j$ denote this model and let $x_{i,-j}$ contain its inputs at operating point $i$. The evaluated reading $y_{ij}$ can be healthy or faulty, and its signed prediction error is $e_{ij}=y_{ij}-h_j(x_{i,-j})$.

To estimate normal prediction error, we divide the healthy observations available for each model fit into five shuffled partitions. Regression models fitted on four partitions predict the readings in the fifth. The population standard deviation of these held-out errors gives a scale $\sigma_j$ for each sensor. We then refit the regression models on all of those healthy observations.

An ensemble of 120 extremely randomized regression trees is also fitted for each sensor, using the same unexpanded inputs~\citep{extratrees}. The population standard deviation of its individual tree predictions at a point is denoted $u_{ij}$. The adjusted absolute prediction error is
\begin{equation}
 z_{ij}=\frac{|e_{ij}|}{\sqrt{\sigma_j^2+u_{ij}^2}+10^{-9}}.
 \label{eq:adjusted_error}
\end{equation}
The regression provides the expected value, while tree disagreement increases the scale when the ensemble gives less consistent predictions. The point's largest adjusted error, $a_i=\max_j z_{ij}$, is compared with the 95th percentile of the corresponding healthy training values. A point is marked suspicious when it exceeds this threshold. To calculate the healthy reference values, we combine the held-out regression errors with tree disagreement from ensembles fitted on all healthy observations available for that fit. Each reference row is therefore excluded from the regression fit that predicts it, while the tree ensemble includes that row.

The operating-condition check separately determines whether speed and torque are close to measured training conditions. After standardizing both variables with training statistics, the mean distance from each training row to its three nearest other training rows is calculated. Its 95th percentile becomes the distance limit. For a new row, the mean distance to the three closest training rows must be no greater than that limit. Only points that are both suspicious and pass this check contribute to the sensor-identification total.

For a group $B$ of $N$ operating points, let $n_B$ count those contributing points. The group raises an alarm when $n_B\geq k$, with $k$ selected from healthy development groups. Values from 1 through $N$ are considered in ascending order. The first whose upper endpoint of a 95\% Wilson interval for the healthy alarm fraction is at most 0.05 is selected~\citep{wilson}. If none meets the rule, $k=N+1$, so this configuration cannot raise an alarm. The healthy groups used in this selection are pooled across the four development partitions. This alarm selection is shared across the sensor-identification comparisons.

The alarm identifies disagreement with the learned healthy relationships. The next step tests which sensor replacement reduces that disagreement.

\section{Identifying the faulty sensor}
\label{app:sensor_identification}
\label{app:methods-source}

An erroneous intake-pressure reading can change the predicted intake-air flow, exhaust-gas-recirculation rate, and nitrogen oxide because those models use pressure as an input. The recorded values of those three measurements can remain healthy while their prediction errors increase. The identification step tests each possible faulty sensor by replacing its reading and measuring the improvement in a separate relationship among measurements~\citep{frank,dunia,qin,alcala}.

For each sensor $c$ under consideration, we first estimate its healthy value from speed, torque, and one other monitored sensor $s$. We check this estimate using a second model that predicts a third sensor $t$. This checking model uses speed, torque, sensor $c$, and a fourth sensor $q$ as inputs. All four sensor roles are distinct, so sensor $s$, which helps estimate the replacement, is neither an input nor the target of the checking model. Both models use the same standardized, second-order ridge specification as the healthy models. Denote the replacement estimate by $\widetilde y_{ic}$ and the checking model by $g_c$.

The checking error before replacement is the absolute difference between the measured target $y_{it}$ and the value predicted by $g_c$. The error after replacement is calculated by inserting $\widetilde y_{ic}$ in place of $y_{ic}$, with the other checking inputs unchanged. For that sensor, the evidence at operating point $i$ is
\begin{equation}
 v_{ic}=
 \frac{|y_{it}-g_c(x_i,y_{ic},y_{iq})|
       -|y_{it}-g_c(x_i,\widetilde y_{ic},y_{iq})|}{b_c},
 \label{eq:replacement_evidence}
\end{equation}
where $x_i$ contains speed and torque. Positive evidence means the replacement reduces the checking error. The denominator $b_c$ is 1.4826 times the median absolute deviation of the checking model's absolute errors on its fitted healthy training rows, plus $10^{-9}$. This allows reductions from different sensor relationships to be compared.

To choose $s$, $t$, and $q$ for each suspected sensor $c$, we compare combinations that satisfy the distinct-sensor requirements above. A useful combination should reduce prediction error when $c$ is faulty while causing little change when its reading is healthy. We evaluate both effects using three shuffled partitions of the observations available for this model fit. For each combination, models fitted on two partitions are applied to the third.

On that third partition, we add a positive offset of two training standard deviations to sensor $c$ and measure the mean reduction in checking error after replacement. We also replace $c$ in the unchanged healthy observations and measure the absolute change in mean checking error. Both quantities are divided by the fitted checking-error scale $b_c$ and averaged across the three partitions. Their difference rewards improvement on faulty readings and penalizes changes on healthy readings.

A combination is eligible when its average fault-related improvement is at least $10^{-6}$. We select the eligible combination with the largest improvement minus penalty, even if that difference is negative, then fit its models and error scale on all observations available for this fit. If no combination qualifies, sensor $c$ is omitted from the available choices.

For each available sensor, we add its replacement-test scores across the contributing points,
\begin{equation}
 V_c(B)=\sum_{i\in B^*}v_{ic},
 \qquad \widehat c_B=\arg\max_c V_c(B),
 \label{eq:group_sensor}
\end{equation}
where $B^*$ contains the suspicious points passing the operating-condition check. The selected sensor is $\widehat c_B$. This selection does not require $n_B\geq k$; a sensor can be selected when fewer contributing points are present than required for an alarm. If no point contributes or no sensor is available, there is no sensor selection, and the associated sensor and selection features are set to zero.

\section{Predicting the fault type}
\label{app:fault_classifier}

A random forest assigns the selected sensor to one of the five fault types~\citep{forest}. Its input combines 12 features describing that sensor and 12 describing the group. The sensor features capture error direction, variation, dependence on the recorded value, and repeated limits or zeros. The group features describe the strength and consistency of the sensor-selection evidence across operating points. The sensor summaries use the sensor selected by the preceding calculation, including when that selection is wrong. Development training therefore incorporates errors carried forward from sensor identification.

To obtain the sensor features, a separate second-order ridge model predicts each sensor using speed and torque alone. For the selected sensor, let $r_i$ be its signed prediction error divided by the population standard deviation of that model's fitted healthy training errors, with $10^{-9}$ added to the scales and denominators. Let $y_i$ be its recorded value and $\ell$ its mean absolute training value. Table~\ref{tab:sensor_features} defines the 12 summaries. These use every point in the group, including points that do not contribute to sensor identification.

\begin{table}[H]
\centering
\caption{Twelve summaries of the sensor selected by the diagnostic procedure. Means, standard deviations, and percentiles are calculated within the group.}
\label{tab:sensor_features}
\begin{tabular}{p{0.34\linewidth}p{0.58\linewidth}}
\toprule
Quantity & Calculation \\
\midrule
Mean absolute error & Mean of $|r_i|$ \\
Mean signed error & Mean of $r_i$ \\
Standard deviation of errors & Standard deviation of $r_i$ \\
Consistency of direction & Absolute mean of $r_i$ divided by its standard deviation plus $10^{-9}$ \\
Error dependence on value & Correlation of $|r_i|$ with $y_i$ \\
Distinct-value fraction & Number of distinct $y_i$, rounded to six decimal places, divided by group size \\
Zero-value fraction & Fraction of $y_i$ equal to zero within numerical tolerance \\
Maximum-value fraction & Fraction equal to the group's largest value \\
Minimum-value fraction & Fraction equal to the group's smallest value \\
Relative mean value & Mean of $y_i$ divided by $\ell+10^{-9}$ \\
Largest absolute error & Maximum of $|r_i|$ \\
Interquartile range of errors & 75th percentile of $r_i$ minus its 25th percentile \\
\bottomrule
\end{tabular}
\end{table}
\FloatBarrier

The correlation is set to zero when fewer than three points are present or either input standard deviation is at most $10^{-9}$. The interquartile range is set to zero for fewer than four points. Equality with the group's extrema allows relative tolerance $10^{-6}$ and absolute tolerance $10^{-8}$; equality with zero uses absolute tolerance $10^{-8}$. We use these conventions in all comparisons.

Group features first describe which operating points contribute to sensor identification. We record the fraction passing the speed-and-torque check, the fraction of those points also exceeding the prediction-error threshold, and the number satisfying both checks. The remaining features use these contributing points to describe how clearly and consistently one sensor is selected.

For each sensor, we sum its replacement-test scores across the contributing points, as in Equation~\ref{eq:group_sensor}. From these totals, we calculate the difference between the largest and second-largest totals and the standard deviation across sensors. Two further inputs divide the largest total by either the sum of all absolute totals or the absolute value of the second-largest total. Both denominators include $10^{-9}$.

At each contributing point, we also select the sensor with the largest replacement-test score. We record the fraction of these selections that match the group's selected sensor and the entropy of their distribution across sensors, using natural logarithms. For each point, we subtract the second-largest score from the largest and record the mean and minimum of these differences across points. Finally, we remove each contributing point in turn and record the fraction of removals that leave the group's selected sensor unchanged.

If no point passes the speed-and-torque check, the fraction exceeding the prediction-error threshold is zero. With just one available sensor, the second-largest score and total are treated as zero. With just one contributing point, removing it is defined to preserve the sensor selection. If no point contributes or no sensor is available, all inputs based on scores and sensor selections are zero. The three inputs describing operating-condition checks and contributing points use their calculated values.

The forest contains 300 trees with balanced class weighting and library defaults for the remaining settings. It is fitted only on development groups in which a known synthetic fault changes at least one reading. Its five probabilities average the tree predictions, and the type with the largest probability is selected. Healthy groups are included in reporting-model training, as described next, but are excluded from fault-type classifier fitting.

\section{Deciding whether to issue a diagnosis}
\label{app:reporting}
\label{app:methods-reporting}

The reporting model estimates whether the diagnosis is correct, following the general use of withheld predictions when evidence is insufficient~\citep{chow,selective}. A development group receives target one only if a fault changes at least one reading, an alarm is raised, at least one point passes the operating-condition check, the selected sensor is correct, and the selected fault type is correct. It receives target zero otherwise. Healthy groups and generated groups with no changed reading therefore have target zero. Replacement accuracy is evaluated separately from diagnosis correctness.

A logistic-regression model predicts this binary target. Its inputs include the 12 group features, the largest fault-type probability, the entropy of the five fault-type probabilities, the difference between their two largest values, and the number of points in the group. The number of contributing points and the fraction passing the speed-and-torque check also appear a second time in the implemented feature vector. This gives 18 inputs from 16 distinct features. Inputs are standardized using the reporting model's training examples. The implementation uses logistic regression with its default squared-coefficient regularization, inverse penalty strength 1.0, and an iteration limit of 2000. If all training targets are equal, a constant model returns that observed fraction.

\begin{samepage}
Threshold selection uses predictions for development groups from models that exclude their partition. For each of the four development partitions, the fitting procedure is
\begin{enumerate}
\item Reserve one partition for evaluation and use only the other three for training the fault-type and reporting models.
\item Within those three training partitions, reserve each one in turn. Fit a fault-type forest on groups with changed fault readings from the other two and predict probabilities for the reserved training partition. Use these probabilities, the fixed development alarm threshold, and the group features to construct the reporting model's training inputs and correctness labels. Fit the reporting model on the three training partitions combined.
\item Fit another fault-type forest on all eligible fault groups from the three training partitions. Predict fault-type probabilities for the evaluation partition, then apply the reporting model to obtain its estimated correctness. Save both the probabilities and correctness estimates for that partition.
\item Repeat for each evaluation partition. The saved correctness estimates determine the reporting threshold, so neither classifier has trained on the groups it scores for threshold selection.
\end{enumerate}
\end{samepage}

For the final reporting fit, we combine all development groups with the fault-type probabilities saved in step 3. The reporting model is fitted on these inputs and their correctness labels. The final fault-type forest is fitted on all development groups in which an injected fault changes a reading. Every forest fit uses 300 trees. Before evaluating the outer test observations, we fit the healthy-prediction and sensor-replacement models on all observations in that split's training portion.

Candidate thresholds are the distinct scores at 60 equally spaced percentile levels from the 0th to the 99th percentile of these development predictions. A threshold is eligible when at least 20 groups receive diagnoses and the lower endpoint of a 95\% Wilson interval for correctness among those diagnoses is at least 0.90. We select the eligible threshold issuing the largest fraction of development diagnoses. If none qualifies, the threshold is set to infinity and that configuration issues no diagnoses. Selection uses the combined healthy and fault development population. The models and thresholds remain fixed when applied to the outer test portion.

For a test group's predicted correctness $q_B$ and selected threshold $\tau$, the report rule is exactly $q_B\geq\tau$. The alarm and operating-condition check contribute features and define correctness, but neither is a separate requirement for issuing a diagnosis. A report issued without an alarm is therefore possible and is counted as incorrect. Alarms and diagnoses issued on healthy groups are measured separately. Withholding a diagnosis leaves its sensor and fault type unreported; it supplies no separate prediction that the measurements are healthy.

All groups from a development partition stay together during these classifier fits. The alarm threshold is selected once from the pooled healthy development groups, while the reporting threshold uses the excluded-partition predictions above. Both criteria use Wilson intervals on generated cases that share measured operating points. They guide threshold selection within this dataset and do not guarantee error rates on independent future observations. The same interpretation applies to the fuel-correction threshold in Appendix~\ref{app:climate}, and Appendix~\ref{app:point-dependence-section} examines sensitivity to the reused operating points.

\section{Diagnosis performance}
\label{app:comparison}

\subsection{Sensor-identification methods}

The sensor-identification comparison keeps the alarm calculation and the speed-and-torque check fixed. It changes the step that identifies the faulty sensor and retrains both the fault-type classifier and the model that decides whether to issue a diagnosis. Every method receives the same generated test groups, so Table~\ref{app:baseline-table} compares the consequences of each identification rule for the complete diagnosis. The reporting and correctness columns describe fault-injected groups, while diagnoses on healthy groups are counted separately in the final column.

Each alternative is fitted within the same training partitions, and its sensor scores are accumulated over the same suspicious points that pass the operating-condition check. The largest-error method uses the adjusted errors in Equation~\ref{eq:adjusted_error}. To account for correlations, we first obtain signed healthy prediction errors from five partitions, with each prediction made by a model fitted on the other four. We standardize these errors and fit a Ledoit--Wolf covariance estimate~\citep{ledoit}. For standardized error vector $r$, fitted mean $\mu$, and inverse covariance $\Omega$, sensor $j$ receives the score $|[(r-\mu)\Omega]_j|/\sqrt{\max(\Omega_{jj},10^{-9})}$.

The principal-component and autoencoder methods reconstruct standardized speed, torque, and all five sensor values from those same seven inputs. Principal-component reconstruction uses three components calculated by full singular-value decomposition. The autoencoder uses hidden layers of 8, 2, and 8 units, hyperbolic-tangent activations, limited-memory quasi-Newton optimization, a regularization parameter of 0.01, and at most 1,000 iterations. Both methods select sensors using absolute reconstruction errors in standardized sensor units. Errors in reconstructed speed and torque are excluded from sensor ranking.

The replacement test and the largest adjusted prediction error give similar overall results. The former issues slightly fewer diagnoses, with slightly greater correctness among those issued. Accounting for correlations sharply reduces diagnoses on healthy groups, while also reducing reporting and correctness on fault groups. These differences show why all three measures are needed to interpret the comparison.

All five variants produce alarms for the same 25,191 of 31,500 fault groups, or 80.0\%, and the same 171 of 2,520 healthy groups, or 6.8\%. Their reporting models can issue a diagnosis without a separate alarm, which explains why the principal-component variant issues more diagnoses on healthy groups than the number of healthy alarms.

\begin{table}[htbp]
\centering\setlength{\tabcolsep}{3pt}
\caption{Comparison of sensor-identification methods using 20 randomly sampled operating points per group. Each row evaluates 31,500 fault groups and 2,520 healthy groups. Sensor identification is measured among fault groups that produce an alarm.}
\label{app:baseline-table}
\begin{tabular}{p{0.35\linewidth}rrrr}
\toprule
Sensor-identification method & \shortstack{Correct sensor\\given alarm} & \shortstack{Diagnoses\\issued} & \shortstack{Correct among\\issued} & \shortstack{Healthy groups\\diagnosed}\\
\midrule
Test estimated replacements & 96.7\% & 77.3\% & 92.1\% & 6.7\%\\
Largest adjusted prediction error & 97.4\% & 78.0\% & 91.8\% & 6.8\%\\
Account for correlation between errors & 82.3\% & 63.4\% & 87.1\% & 0.2\%\\
Principal-component reconstruction & 91.8\% & 74.6\% & 88.8\% & 11.5\%\\
Autoencoder reconstruction & 85.7\% & 63.1\% & 91.8\% & 3.0\%\\
\bottomrule
\end{tabular}
\end{table}
\FloatBarrier

The autoencoder did not meet its convergence criterion in 19 of the 21 combinations of training split and random seed. Its row therefore describes the fitted architecture under these settings, including runs that stopped before convergence.

\FloatBarrier
\subsection{Reporting threshold}

\begin{figure}[htbp]
\centering
\includegraphics[width=\linewidth]{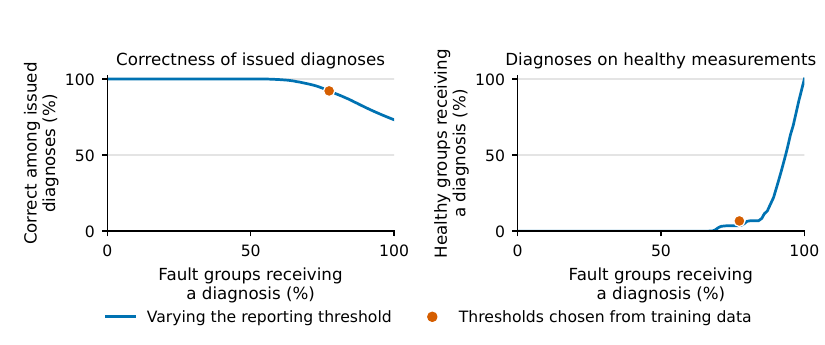}
\caption{Diagnosis as the reporting threshold changes, calculated from saved predictions for 31,500 fault groups and 2,520 healthy groups. The highlighted points show the thresholds selected during training, placing the reported results on these curves.}
\label{fig:reporting}
\end{figure}

The reporting threshold determines how often the system issues a diagnosis. Figure~\ref{fig:reporting} shows the associated change in correctness among issued fault diagnoses and in diagnoses issued on healthy measurements. The highlighted decisions use training-selected thresholds. The remaining test curves describe the consequences of varying a threshold and are not used to choose it.

\subsection{Fault types and their relative frequencies}

Table~\ref{app:fault-types} gives exact counts for the five fault types. The pooled result gives more weight to offsets, which contribute three tested magnitudes in two directions, than to clipping, which contributes two limits. Differences across fault types are substantial. Almost every issued noise diagnosis is correct, while correctness is appreciably lower for scaling errors and clipping.

\begin{table}[htbp]
\centering\setlength{\tabcolsep}{4pt}
\caption{Results by fault type for 20 randomly sampled operating points. Correct issued diagnoses require a changed reading, an alarm, an accepted operating point, and correct identification of both the sensor and fault type.}
\label{app:fault-types}
\begin{tabular}{lrrrrr}
\toprule
Fault type & Groups & \shortstack{Alarm\\raised} & \shortstack{Diagnoses\\issued} & \shortstack{Correct\\issued} & \shortstack{Correct among\\issued}\\
\midrule
Added offset & 9,450 & 99.9\% & 9,123 & 8,799 & 96.4\%\\
Scaling error & 6,300 & 57.1\% & 2,931 & 1,849 & 63.1\%\\
Added noise & 6,300 & 100.0\% & 6,287 & 6,286 & 99.98\%\\
Intermittent zero readings & 6,300 & 80.1\% & 5,155 & 4,948 & 96.0\%\\
Clipping & 3,150 & 25.7\% & 868 & 566 & 65.2\%\\
\midrule
All generated groups & 31,500 & 80.0\% & 24,364 & 22,448 & 92.1\%\\
\bottomrule
\end{tabular}
\end{table}
\FloatBarrier

The corresponding reporting rate is 77.3\%. Restricting evaluation to the 30,574 groups in which at least one reading actually changes gives an alarm rate of 82.3\%, a reporting rate of 79.6\%, and 92.3\% correctness among issued diagnoses. This restriction principally changes the denominators for clipping and intermittent zero readings. All generated injections contribute to the main result, including those that leave readings unchanged.

To evaluate an equally likely choice of fault type, let $N_m$ be the number of evaluated groups of type $m$, $A_m$ the number receiving diagnoses, and $C_m$ the number of correct issued diagnoses. The reporting rate and correctness among issued diagnoses under this assumption are
\begin{equation}
 R_{\mathrm{equal}}=\frac{1}{5}\sum_{m=1}^{5}\frac{A_m}{N_m},
 \qquad
 P_{\mathrm{equal}}=\frac{\sum_{m=1}^{5} C_m/N_m}{\sum_{m=1}^{5} A_m/N_m}.
\label{app:equal-weighting}
\end{equation}
They are 70.4\% and 90.5\%, respectively, with a 72.6\% alarm rate. Directly averaging the five conditional correctness percentages gives 84.1\%. That average answers a different question because it gives equal weight to a fault type even when the system rarely issues a diagnosis for it.

\subsection{Group size and operating conditions}

Table~\ref{app:group-size} compares groups of 5, 10, and 20 distinct operating points. Random selection samples without replacement from the test split. The nearby-point construction selects an initial point and fills the group with its closest points in standardized speed and torque. Each construction uses its own thresholds, selected during training and fixed during testing.

\begin{table}[htbp]
\centering\setlength{\tabcolsep}{4pt}
\caption{Group-size experiment. Each row contains 31,500 fault groups and 2,520 healthy groups. The two healthy columns distinguish alarms from issued diagnoses.}
\label{app:group-size}
\begin{tabular}{llrrrrr}
\toprule
Points & Construction & \shortstack{Fault\\alarms} & \shortstack{Diagnoses\\issued} & \shortstack{Correct among\\issued} & \shortstack{Healthy\\alarms} & \shortstack{Healthy groups\\diagnosed}\\
\midrule
5 & Random & 68.8\% & 62.7\% & 94.2\% & 1.6\% & 0.0\%\\
5 & Nearby & 62.4\% & 54.9\% & 94.0\% & 1.5\% & 0.0\%\\
10 & Random & 75.1\% & 70.9\% & 93.0\% & 3.1\% & 0.2\%\\
10 & Nearby & 71.3\% & 64.8\% & 92.1\% & 3.6\% & 0.3\%\\
20 & Random & 80.0\% & 77.3\% & 92.1\% & 6.8\% & 6.7\%\\
20 & Nearby & 78.6\% & 73.5\% & 92.7\% & 5.1\% & 5.1\%\\
\bottomrule
\end{tabular}
\end{table}
\FloatBarrier

Larger groups increase the fraction of faults producing alarms and diagnoses, while increasing healthy-data errors in this experiment. Randomly selected groups receive more diagnoses at every tested size. Correctness among issued diagnoses does not consistently favor either construction.

\subsection{Diagnoses on healthy measurements}

All 169 diagnoses issued for healthy 20-point random groups arise in the test split withholding 550, 1,200, and 1,400 revolutions per minute. Seed 42 issues 90 diagnoses among that split's 120 healthy groups, and seed 123 issues 79 among 120. Their alarm counts are 90 and 81, respectively. Seed 2026 issues no diagnoses on healthy groups in this split, and the other 18 split-and-seed combinations also issue none. Thus, the 6.7\% overall rate combines concentrated failure in two settings with zero observed errors in the remaining 19.

\subsection{Fault magnitude and sensor differences}

The magnitude experiment freezes the fitted models and thresholds while varying offsets, scaling errors, and clipping limits. An offset is measured in units of the sensor's healthy training standard deviation. A scaling error is the fractional change in its recorded value. Clipping limits are percentiles of the healthy training readings, so higher percentiles usually produce weaker faults and can leave some groups entirely unchanged.

Table~\ref{app:sensor-severity} illustrates the sensor differences using three settings. Each offset and scaling entry combines positive and negative directions over 630 groups per sensor. Small offsets can produce alarms while the resulting sensor-and-fault diagnoses remain incorrect. Increasing offset magnitude improves correctness among issued diagnoses for all five sensors. An 8\% scaling error remains difficult for several sensors, including intake-air mass flow, for which none of the 103 issued diagnoses in this setting is completely correct. The pooled result therefore hides differences across sensors and fault magnitudes.

\begin{table}[htbp]
\centering\setlength{\tabcolsep}{3pt}
\caption{Fraction of groups receiving diagnoses and correctness among those diagnoses in selected magnitude tests. Each cell gives the two percentages in that order. Offset magnitudes are in healthy training standard deviations.}
\label{app:sensor-severity}
\begin{tabular}{p{0.40\linewidth}rrr}
\toprule
Faulty sensor & \shortstack{Offset 0.25\\issued / correct} & \shortstack{Offset 3.0\\issued / correct} & \shortstack{Scaling 8\%\\issued / correct}\\
\midrule
Intake-manifold temperature & 25.9 / 29.4 & 100.0 / 100.0 & 63.8 / 37.8\\
Intake-manifold pressure & 60.0 / 38.4 & 100.0 / 100.0 & 30.5 / 77.1\\
Exhaust-gas-recirculation rate & 14.6 / 17.4 & 100.0 / 100.0 & 17.9 / 27.4\\
Intake-air mass flow & 36.5 / 10.4 & 100.0 / 100.0 & 16.3 / 0.0\\
Engine-out nitrogen oxide & 10.6 / 29.9 & 96.8 / 98.5 & 4.6 / 79.3\\
\bottomrule
\end{tabular}
\end{table}
\FloatBarrier

\begin{figure}[htbp]
\centering
\includegraphics[width=\linewidth]{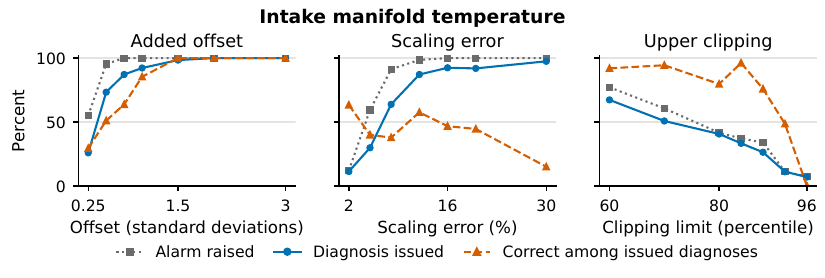}
\caption{Intake-manifold temperature diagnosis as fault magnitude changes, pooling three seeds and seven test splits with fixed models and thresholds. The curves give alarm frequency, diagnosis frequency, and correctness among issued diagnoses, using the units in Figure~\ref{fig:severity}.}
\label{fig:sensor-temperature}
\end{figure}

\begin{figure}[htbp]
\centering
\includegraphics[width=\linewidth]{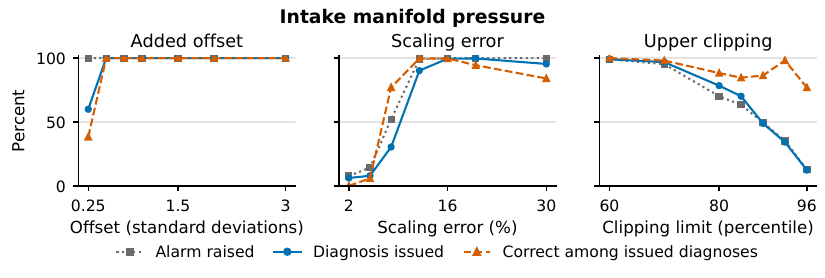}
\caption{Intake-manifold pressure diagnosis as fault magnitude changes, pooling three seeds and seven test splits with fixed models and thresholds. The curves give alarm frequency, diagnosis frequency, and correctness among issued diagnoses, using the units in Figure~\ref{fig:severity}.}
\label{fig:sensor-pressure}
\end{figure}

\begin{figure}[htbp]
\centering
\includegraphics[width=\linewidth]{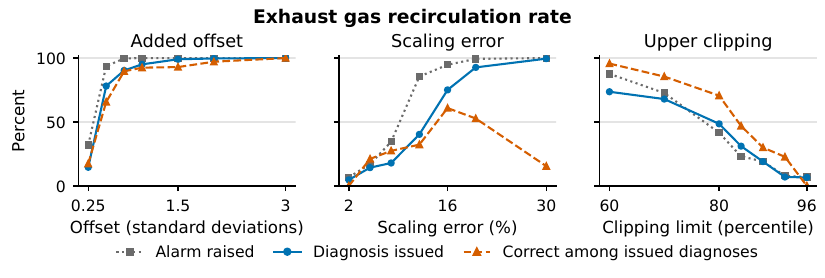}
\caption{Exhaust-gas-recirculation rate diagnosis as fault magnitude changes, pooling three seeds and seven test splits with fixed models and thresholds. The curves give alarm frequency, diagnosis frequency, and correctness among issued diagnoses, using the units in Figure~\ref{fig:severity}.}
\label{fig:sensor-recirculation}
\end{figure}

\begin{figure}[htbp]
\centering
\includegraphics[width=\linewidth]{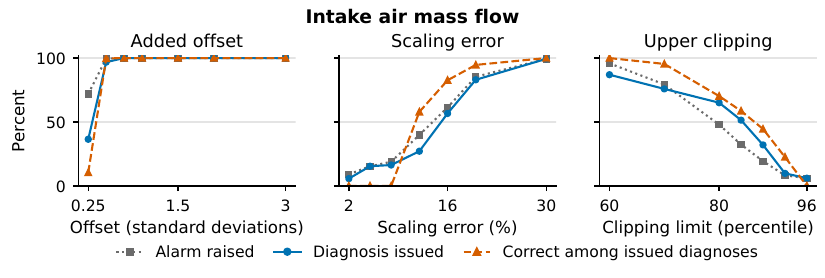}
\caption{Intake-air mass flow diagnosis as fault magnitude changes, pooling three seeds and seven test splits with fixed models and thresholds. The curves give alarm frequency, diagnosis frequency, and correctness among issued diagnoses, using the units in Figure~\ref{fig:severity}.}
\label{fig:sensor-air}
\end{figure}

\begin{figure}[htbp]
\centering
\includegraphics[width=\linewidth]{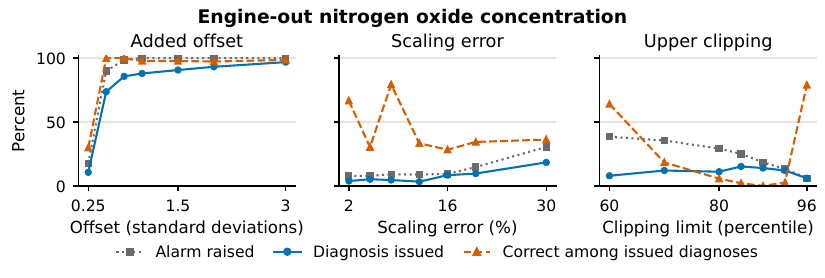}
\caption{Engine-out nitrogen oxide concentration diagnosis as fault magnitude changes, pooling three seeds and seven test splits with fixed models and thresholds. The curves give alarm frequency, diagnosis frequency, and correctness among issued diagnoses, using the units in Figure~\ref{fig:severity}.}
\label{fig:sensor-nitrogen}
\end{figure}

Figures~\ref{fig:sensor-temperature}--\ref{fig:sensor-nitrogen} give the complete sensor-specific magnitude curves. They use the same frozen models and thresholds as the pooled plot in Figure~\ref{fig:severity}.

\FloatBarrier
\subsection{Faults outside the training categories}

Three additional tests apply load-dependent offsets, stuck readings, and simultaneous offsets to two sensors. The load-dependent offset equals the sensor's training standard deviation multiplied by standardized torque, with both signs tested. A stuck reading is fixed at the training median plus or minus one training standard deviation. For each of the ten sensor pairs, the simultaneous-fault test adds 1.5 training standard deviations to one sensor and subtracts 1.5 from the other. The fitted five-type classifier and reporting rule remain fixed.

All 3,150 groups in each test produce alarms. Nevertheless, the system issues a familiar fault label for 3,103 load-dependent-offset groups, 2,840 stuck-reading groups, and 3,094 two-sensor groups. These are reporting rates of 98.5\%, 90.2\%, and 98.2\%. The classifier has no output representing these additional fault families, so the experiment measures how frequently the system gives a familiar diagnosis outside its training categories. Withholding uncertain diagnoses does not reliably distinguish these unfamiliar faults.

\FloatBarrier
\section{Component comparisons and sensitivity analyses}
\label{app:components}
\label{app:earlier}

The component-design comparison uses five randomly divided observation folds with seed 42 and seven folds that withhold all measurements at selected engine speeds. Each observation is tested once in each evaluation setting. The random folds can contain the same speed in training and testing, so they measure performance among conditions represented elsewhere on the recorded engine map. Each engine-speed fold instead excludes all measurements at three speeds. Healthy models and numerical thresholds are fitted using each fold's training observations. These comparisons informed the architecture and assess the developed design on the same engine map used for model selection.

For this component comparison, the fault-type and reporting models are fitted using fault groups only. During classifier fitting, we keep together all generated groups sharing an outer fold, injected sensor, fault type, magnitude, direction, and random realization. A first prediction pass obtains fault-type probabilities from classifiers fitted without the set of groups being evaluated. A second pass uses those probabilities to obtain reporting scores from models estimating diagnosis correctness, again fitted without that set of groups. Repeated fault-type fits contain 200 trees, and the final fault-type classifier contains 300.

The component comparison counts a diagnosis as correct when an alarm is raised, at least one operating point passes the condition check, and both the sensor and fault type are identified correctly. It does not separately require the injection to change a reading. The reporting threshold requires at least 20 development diagnoses and the lower endpoint of a 95\% Wilson interval of at least 90\% correctness, and is fixed before testing.

The main diagnosis experiment uses the four development partitions described in Appendix~\ref{app:splits}. All groups from the partition being evaluated remain together, while its fault-type and reporting models are fitted on groups from the other three partitions. The reporting model's training inputs are generated through the further fitting steps in Appendix~\ref{app:methods-reporting}. Healthy groups and injections that leave readings unchanged are included as incorrect examples, correctness requires a changed reading, and every fault-type fit uses 300 trees. The component and main results therefore evaluate different classifier-training procedures and correctness definitions.

\subsection{Healthy prediction models}

The healthy-model comparison uses five shuffled observation folds and predicts each of five sensors from speed, torque, and the other four sensors. It measures healthy prediction accuracy and sensor identification under a positive offset of two training standard deviations. Table~\ref{app:legacy-models} shows that the best healthy fit does not automatically produce the best identification result. Polynomial ridge regression was selected using both measures.

\begin{table}[htbp]
\centering\caption{Healthy-model component comparison, averaged across sensors and folds. Sensor identification uses one positive-offset setting.}
\label{app:legacy-models}
\begin{tabular}{lrr}
\toprule
Model & Coefficient of determination & Correct sensor\\
\midrule
Linear ridge regression & 0.9291 & 80.6\%\\
Second-order polynomial ridge regression & 0.9910 & 92.6\%\\
Partial least squares regression & 0.9086 & 68.7\%\\
Extremely randomized trees & 0.9924 & 78.2\%\\
Random forest & 0.9807 & 64.7\%\\
Five-neighbour regression & 0.9769 & 75.4\%\\
\bottomrule
\end{tabular}
\end{table}
\FloatBarrier

\subsection{Prediction-error scaling and operating-condition checks}

Alarm comparisons at individual operating points informed the use of tree disagreement. Table~\ref{app:legacy-adjustment} keeps these component results separate from the later group-level alarms. A second comparison found that 63.1\% of points passed an operating-condition check requiring both location within the boundary of training speed-and-torque points and proximity to training observations. Among those points, 4.1\% produced healthy alarms. The fractions passing the other checks and their healthy alarm rates were 90.3\% and 16.7\% for isolation forest, 82.5\% and 9.4\% for kernel density, 70.5\% and 5.8\% for nearest neighbours, and 90.8\% and 15.6\% for covariance-based distance. All started from the same 17.5\% alarm rate before checking the operating condition. Alarm rates after checking the operating condition are calculated only among points that pass the check, so each rate should be interpreted alongside that fraction.

\begin{table}[htbp]
\centering\caption{Alarm rates on healthy operating points under different adjustments of prediction error.}
\label{app:legacy-adjustment}
\begin{tabular}{lrr}
\toprule
Adjustment & Random point split & Held-out speeds\\
\midrule
Global healthy error scale & 8.7\% & 20.7\%\\
Speed-and-load region scale & 6.9\% & 17.5\%\\
Disagreement between randomized trees & 1.4\% & 3.7\%\\
Nearby-point error scale & 5.5\% & 19.4\%\\
\bottomrule
\end{tabular}
\end{table}
\FloatBarrier

\subsection{Sensor identification and fault classification}

The component comparison selected one sensor for each of 5,806 changed readings, without withholding any selection. Correct sensor identification was 63.0\% for principal-component errors, 83.6\% for the largest adjusted prediction error, 84.7\% after accounting for correlation among errors, and 88.3\% for testing estimated replacements with a separate checking model. These values measure individual sensor choices, whereas Table~\ref{app:baseline-table} evaluates the complete procedure across three seeds.

A fault-type component experiment achieved 42.8\% accuracy with one operating point, 81.1\% with five, 87.7\% with ten, and 92.8\% with twenty. These values measure fault-type classification alone and do not require correct sensor identification or a correct complete diagnosis. A separate two-sensor test applied positive offsets of two training standard deviations to five sensor pairs over the full 217-point map using randomly divided observations. The first selected sensor belonged to the faulty pair in 98.7\% of cases. Repeating identification after replacement recovered both sensors in 86.1\%. Substantial disagreement remained after the first replacement in 99.9\% of two-fault cases, compared with 15.2\% of healthy cases. This test examined repeated sensor selection under one offset setting and did not evaluate complete two-sensor fault-type diagnoses.

\subsection{Alarm and reporting components}

Table~\ref{app:legacy-architecture} compares combinations of alarm scaling and reporting rules. Regional error adjustment produced no issued diagnoses satisfying the development reporting criterion. Tree-disagreement adjustment enabled reporting, and using a model estimating diagnosis correctness increased the reporting rate further. The final two rows give results for this reporting model under the random and engine-speed splits.

\begin{table}[htbp]
\centering\setlength{\tabcolsep}{3pt}
\caption{Single-seed 20-point component comparison. Random point splits contain 7,500 fault groups, and held-out-speed splits contain 10,500. The reporting rule requires a training Wilson interval lower endpoint of 90\% for correctness among issued diagnoses.}
\label{app:legacy-architecture}
\begin{tabular}{L{0.30\linewidth}lrrrr}
\toprule
Design and reporting rule & Split & \shortstack{Healthy\\alarms} & \shortstack{Fault\\alarms} & \shortstack{Diagnoses\\issued} & \shortstack{Correct among\\issued}\\
\midrule
Regional adjustment & Random & 0.0\% & 78.1\% & 0.0\% & Undefined\\
\addlinespace[4pt]
Regional adjustment & Held-out speed & 14.3\% & 77.6\% & 0.0\% & Undefined\\
\addlinespace[4pt]
\makecell[l]{Tree adjustment\\Fault-type probability} & Random & 0.0\% & 82.1\% & 70.8\% & 97.9\%\\
\addlinespace[4pt]
\makecell[l]{Tree adjustment\\Fault-type probability} & Held-out speed & 10.7\% & 79.5\% & 62.6\% & 97.5\%\\
\addlinespace[4pt]
\makecell[l]{Tree adjustment\\Diagnosis correctness} & Random & 0.0\% & 82.1\% & 75.3\% & 96.7\%\\
\addlinespace[4pt]
\makecell[l]{Tree adjustment\\Diagnosis correctness} & Held-out speed & 10.7\% & 79.5\% & 72.5\% & 94.6\%\\
\bottomrule
\end{tabular}
\end{table}
\FloatBarrier

Before withholding, diagnosis correctness over all groups increased from 74.9\% to 77.0\% with tree adjustment in the random split, and from 71.3\% to 72.6\% at held-out speeds. Correct sensor identification among alarms changed from 99.4\% to 98.3\%, and from 97.6\% to 96.6\%, respectively. Table~\ref{app:legacy-fault-types} gives the fault-type breakdown for this single-seed experiment. Its arithmetic averages across the five rows are 72.1\% for alarms, 90.0\% for correct sensor identification among alarms, 65.2\% for reporting, and 87.0\% for correctness among issued diagnoses. These are per-type averages, and the last two conditional averages differ from the equal-fault-type calculation in Equation~\ref{app:equal-weighting}.

\begin{table}[htbp]
\centering\setlength{\tabcolsep}{4pt}
\caption{Single-seed held-out-speed results by fault type for 20-point groups. These results use 10,500 generated groups and different reporting training and correctness definitions from the three-seed results in Table~\ref{app:fault-types}.}
\label{app:legacy-fault-types}
\begin{tabular}{lrrrrr}
\toprule
Fault type & Groups & \shortstack{Alarm\\raised} & \shortstack{Correct sensor\\given alarm} & \shortstack{Diagnoses\\issued} & \shortstack{Correct among\\issued}\\
\midrule
Added offset & 3,150 & 99.7\% & 99.9\% & 93.9\% & 96.5\%\\
Scaling error & 2,100 & 56.3\% & 84.6\% & 31.4\% & 72.1\%\\
Added noise & 2,100 & 100.0\% & 100.0\% & 99.5\% & 99.9\%\\
Intermittent zero readings & 2,100 & 78.9\% & 99.5\% & 79.7\% & 96.8\%\\
Clipping & 1,050 & 25.3\% & 65.8\% & 21.6\% & 70.0\%\\
\bottomrule
\end{tabular}
\end{table}
\FloatBarrier

\subsection{Offset detection at individual operating points}

\begin{figure}[htbp]
\centering
\includegraphics[width=\linewidth]{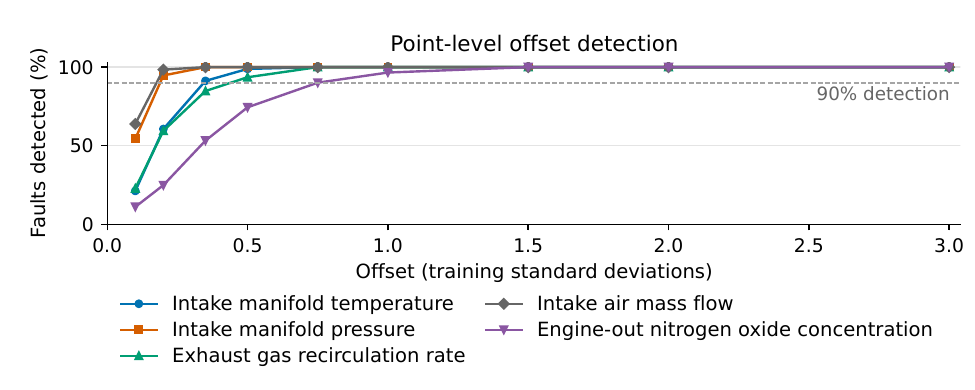}
\caption{The curves show how often individual operating points raise alarms as offset magnitude increases, using prediction errors scaled within regions of the engine map. Each sensor and magnitude includes 434 positive and negative offset injections across five shuffled folds with seed 42, with the dashed line marking 90\% detection.}
\label{fig:point-offsets}
\end{figure}

Figure~\ref{fig:point-offsets} shows detection across the complete tested offset grid in the component experiment at individual operating points. The grid contains 0.10, 0.20, 0.35, 0.50, 0.75, 1.0, 1.5, 2.0, and 3.0 healthy training standard deviations. Each sensor and magnitude uses 434 injections, combining both offset directions over the 217 physical points. This analysis measures detection at individual operating points. The complete-diagnosis magnitude experiment in Figure~\ref{fig:severity} additionally requires a correct sensor and fault type.

The smallest tested magnitude reaching 90\% detection differs across sensors. These magnitudes were 0.35 training standard deviations for intake temperature, 0.20 for intake pressure, 0.50 for exhaust-gas recirculation, 0.20 for intake-air flow, and 0.75 for nitrogen oxide. Their median physical sizes were 3.23 degrees Celsius, 0.155 bar, 4.07 percentage points, 74.95 kilograms per hour, and 238.43 parts per million.

\subsection{Sensitivity to removing measured operating points}
\label{app:point-dependence-section}

Repeated synthetic faults and group construction reuse the same 217 diesel operating points. The single-seed sensitivity analysis removes each physical point in turn, discards every group containing it, and recalculates performance with the fitted models fixed. If $\widehat\theta_{(-p)}$ is the resulting measure and $\overline\theta_{(-)}$ its mean over the $M=217$ deletions, the standard-error estimate is
\begin{equation}
 s_{\mathrm{delete}}=
 \sqrt{\frac{M-1}{M}\sum_{p=1}^{M}
 \left(\widehat\theta_{(-p)}-\overline\theta_{(-)}\right)^2}.
\label{app:delete-point}
\end{equation}
The interval extends 1.96 such standard errors around the full-data estimate, with endpoints limited to zero and one. This calculation measures sensitivity to point membership while keeping learned models fixed.

\begin{table}[htbp]
\centering\caption{Single-seed held-out-speed results and delete-one-point sensitivity intervals. These intervals apply to the single-seed generated groups and are not uncertainty intervals for the three-seed results.}
\label{app:point-dependence}
\begin{tabular}{lrrrr}
\toprule
Measure & \shortstack{5-point\\estimate} & \shortstack{5-point\\interval} & \shortstack{20-point\\estimate} & \shortstack{20-point\\interval}\\
\midrule
Fault alarms & 67.8\% & 61.6--74.1\% & 79.5\% & 55.5--100.0\%\\
Correct sensor among alarms & 97.1\% & 94.3--100.0\% & 96.6\% & 82.8--100.0\%\\
Diagnoses issued & 59.0\% & 53.4--64.7\% & 72.5\% & 52.5--92.4\%\\
Correct among issued diagnoses & 96.3\% & 95.0--97.7\% & 94.6\% & 86.0--100.0\%\\
\bottomrule
\end{tabular}
\end{table}
\FloatBarrier

Deleting one point removes many overlapping 20-point groups, producing wider intervals than for five-point groups. The three-seed experiment increases variation in random generation and fitted models through three seeds, while reusing the same physical map. Its larger group count consequently supplies more generated tests without adding independent engine measurements.

\FloatBarrier
\section{Fuel correction and carbon dioxide accounting}
\label{app:climate}

\subsection{Fuel-correction procedure}

The accounting experiment tests faults in all six sensors and includes healthy scenarios during development and testing. It evaluates each operating point and replaces a fuel reading only when fuel flow is selected as faulty. To select a sensor, the procedure estimates each sensor's healthy value from speed, torque, and the other five readings. It inserts that estimate and recalculates the largest adjusted prediction error across the six sensors. The reduction divided by the original largest error plus $10^{-9}$ gives that sensor's score.

Fuel correction requires an alarm, speed and torque passing the operating-condition check, selection of fuel flow, and a sufficient difference between the two largest sensor scores. This difference determines whether the sensor choice is accepted. Here, every replacement estimate uses speed, torque, and all five other sensor readings. The primary diagnosis instead estimates a replacement using speed, torque, and one other sensor, then checks its effect on a different sensor with a separate model. The fuel-correction procedure also uses its own acceptance rule and does not use the primary diagnosis experiment's fault-type classifier or reporting model.

The six-sensor monitor uses standardized second-order ridge models with penalty 1.0. Healthy prediction errors are obtained from five partitions that keep complete engine-speed levels together. Their standard deviations and disagreement among 120 randomized trees scale the errors as in Equation~\ref{eq:adjusted_error}. The alarm threshold for individual operating points is the 95th percentile of the largest adjusted error on healthy fitting observations. Tree disagreement is computed from ensembles fitted on all healthy observations available for that model fit. The operating-condition check uses the three-neighbour rule in Appendix~\ref{app:alarm}. Development faults use the strengths in Table~\ref{tab:fault_grid} and one realization for noise and intermittent zeros. Test faults use three realizations.

Within each outer training split, four development partitions provide readings evaluated by models fitted on the other three. Development includes healthy readings, faults in fuel flow, and faults in the other five sensors. We consider a fuel reading for correction when it raises an alarm, passes the operating-condition check, and is selected as faulty. That selection is labelled correct only when the injected fault is in fuel flow and changes the reading.

We calculate candidate thresholds from the observed differences between the two largest sensor scores, using 81 equally spaced percentile levels from the minimum to the maximum. An eligible threshold accepts at least 25 fuel selections from at least 10 distinct measured operating points, with a lower endpoint of a 95\% Wilson interval of at least 90\% correct fuel selections. The eligible threshold accepting the most selections is fixed for testing. If none qualifies, the threshold is infinite and no fuel reading is corrected. The shared interpretation of this threshold-selection criterion is given in Appendix~\ref{app:methods-reporting}.

The evaluation uses seven splits that exclude all measurements at selected engine speeds and seeds 42, 123, and 2026. Development magnitudes differ from final test magnitudes. The monitor excludes brake-specific fuel consumption and injected fuel quantity because brake-specific consumption is calculated from fuel use and power, while injected fuel quantity directly describes fuel delivery. Ambient temperature and pressure are also excluded from this experiment. The input set therefore contains speed, torque, and the six separately measured sensor quantities. Each reconstructed fuel value is calculated from speed, torque, and the other five sensor readings, without the corrupted fuel reading as an input.

\subsection{Accounting assumptions and error measures}

Let $f_i$ be the healthy fuel rate in kilograms per hour and $f_i^{(a)}$ the final rate under treatment $a$. Let $w_i$ be assigned operating time in hours, with $\sum_i w_i=1$. The calculated combustion carbon dioxide and its error in one scenario are
\begin{equation}
 E^{(a)}=\kappa\sum_i w_i f_i^{(a)},\qquad
 E^{\mathrm{healthy}}=\kappa\sum_i w_i f_i,\qquad
 L^{(a)}=\left|E^{(a)}-E^{\mathrm{healthy}}\right|.
\label{app:co2-scenario}
\end{equation}
The default factor is $\kappa=3.1863$ kilograms of carbon dioxide per kilogram of fossil diesel. It follows from the Intergovernmental Panel on Climate Change defaults of 43 terajoules per gigagram of fuel and 74,100 kilograms of carbon dioxide per terajoule \citep{ipcc}. Fuel composition can change this factor.

Equal-time weights assign $1/217$ hours to each physical operating point. For the other schedules, torque is scaled to $q_i$ between zero and one. Low-load weights are proportional to $1+4(1-q_i)$, and high-load weights to $1+4q_i$. Both sets are normalized to one hour. These durations are assumed because the dataset records steady-state conditions without driving-cycle durations or a temporal sequence. Some synthetic injections produce negative fuel readings; these remain in the accounting calculation when the correction rule leaves them unchanged.

The absolute error in Equation~\ref{app:co2-scenario} is calculated before averaging scenarios. Consequently, an overestimate in one scenario cannot cancel an underestimate in another. Within faulty scenarios, the five fault types receive equal weight, and their tested magnitudes, directions, and random realizations receive equal weight within each type. The sensors included in each comparison receive equal fault probability. If $\pi$ is the share of one-hour scenarios containing a fault, the mean absolute error is
\begin{equation}
 \overline L^{(a)}(\pi)
 =(1-\pi)\overline L^{(a)}_{\mathrm{healthy}}
 +\pi\overline L^{(a)}_{\mathrm{fault}}.
\label{app:fault-hour-mixture}
\end{equation}
The main assumption $\pi=0.01$ therefore assigns 99\% weight to healthy one-hour scenarios and 1\% to faulty ones. It does not say that 1\% of the points in every hour have a fault. Separate saved calculations use fault shares of 0\%, 0.1\%, 1\%, 5\%, and 10\%, and consider faults in fuel flow alone, the other five sensors, or all six sensors with equal probability.

\subsection{Accounting results}

\begin{table}[htbp]
\centering\setlength{\tabcolsep}{4pt}
\caption{Mean absolute error in calculated combustion carbon dioxide, in kilograms per engine-hour. Results assign 1\% of scenario weight to faulty hours, with equal weights for six sensors and five fault types, and average across three seeds.}
\label{app:accounting-schedules}
\begin{tabular}{lrrr}
\toprule
Fuel-reading treatment & Equal time & Low-load weighting & High-load weighting\\
\midrule
Use recorded readings & 0.052908 & 0.050342 & 0.056073\\
Correct selected readings & 0.008552 & 0.008437 & 0.009058\\
Replace every reading & 0.348838 & 0.423068 & 0.260979\\
Correct readings known to be faulty & 0.000325 & 0.000399 & 0.000237\\
\bottomrule
\end{tabular}
\end{table}
\FloatBarrier

The reduction from using recorded readings to correcting selected readings is 83.8\% with equal time, 83.2\% with low-load weighting, and 83.8\% with high-load weighting. The comparison using the recorded injected faults corrects only fuel readings changed by those faults. It uses the same estimated replacement values, so their prediction errors remain in the calculation.

Across all accounting test records, the rule corrects 7,803 of 11,591 fuel readings changed by injected faults, or 67.3\%. No correction is applied to any of the 82,804 unchanged fuel readings, including those recorded while another sensor is faulty. None of the selected replacements increases absolute fuel error.

The total-error measure allows errors at different operating points to cancel within a single hour. As a complementary measure, the time-weighted absolute pointwise error is
\begin{equation}
 L_{\mathrm{point}}^{(a)}=\kappa\sum_i w_i\left|f_i^{(a)}-f_i\right|.
\label{app:pointwise-accounting}
\end{equation}
Using equal time and 1\% faulty scenarios with equal fault probability across six sensors gives mean values of 0.087096 kilograms for recorded readings, 0.013719 for selected corrections, 1.631486 for replacing every fuel reading, and 0.001771 for correction of readings known to be faulty. Selected correction reduces this measure by 84.2\%. This measure calculates absolute error at each operating point before summing, so positive and negative errors cannot cancel as they can in the one-hour total. Its values therefore differ from Table~\ref{app:accounting-schedules}.

\subsection{Fuel-only reconstruction experiment}
\label{app:fuel-reconstruction}

The fuel-only reconstruction experiment applies faults to fuel flow alone while allowing any of the six sensors to be selected as faulty. It evaluates individual changed fuel readings and reports the selected sensor without predicting a fault type. As in the accounting experiment, replacement fuel values are estimated from speed, torque, and the other five sensor readings. Table~\ref{app:legacy-fuel} gives the results.

To choose the sensor-selection threshold, faults are added to fuel readings within each training split. Only changed readings that raise an alarm and pass the operating-condition check contribute to this choice. Candidate thresholds are taken at 80 quantiles of the differences between the two largest sensor scores, from the zeroth through the 98th percentile. Each qualifying threshold must accept at least ten sensor selections and give a lower endpoint of a 95\% Wilson interval of at least 90\% correct fuel identification. The threshold accepting the most selections is then fixed for testing.

If no candidate meets these requirements, sensor selections are withheld. When fewer than ten score differences are available to choose a threshold, however, the implementation allows every selection that passes the remaining checks. The accounting experiment withholds fuel corrections when its threshold-selection requirements cannot be met.

The fuel-only implementation also restricts correction to readings marked as changed by the injected fault, and all readings in its reported test population satisfy that condition. The accounting experiment instead includes healthy readings and faults in all six sensors. Its correction decisions use no injected-fault record; those records are used only to choose the threshold during training and to evaluate the results.

\begin{table}[H]
\centering\setlength{\tabcolsep}{3pt}
\caption{Fuel-only experiment at individual operating points, restricted to changed fuel readings. Error removal measures summed absolute fuel-reading error.}
\label{app:legacy-fuel}
\begin{tabular}{lrrrrrr}
\toprule
Split & \shortstack{Changed\\points} & \shortstack{Fault\\alarms} & \shortstack{Selection\\reported} & \shortstack{Fuel correct\\among reports} & \shortstack{Error\\removed} & \shortstack{Error removed\\known faults}\\
\midrule
Random & 2,765 & 70.2\% & 66.3\% & 100.0\% & 89.7\% & 98.2\%\\
Held-out speed & 2,778 & 71.2\% & 63.2\% & 99.9\% & 83.8\% & 98.0\%\\
\bottomrule
\end{tabular}
\end{table}
\FloatBarrier

The fuel-reading error-removal measure is one minus the sum of final absolute fuel errors divided by the sum of corrupted absolute fuel errors. Missed and withheld corrections remain in the sum. The comparison using the injected-fault records applies the same estimated replacement to every changed fuel reading.

Correct sensor selection among alarms was 99.6\% and 99.8\%, respectively. No increase in absolute fuel error was observed after a selected correction in either split. The 83.8\% error-removal figure measures summed absolute fuel-reading error over changed points. The equal 83.8\% reduction in Table~\ref{app:accounting-schedules} measures mean absolute error in a one-hour carbon dioxide total, including healthy scenarios and faults in other sensors. The percentages refer to different error measures and evaluation populations.

\clearpage
\section{Turbofan sensor detection and reconstruction}
\label{app:turbofan}

The Turbofan Sensor-FDI-Bench supplies synthetic steady-state measurements from a physics-based turbofan model, including clean references and structured sensor faults \citep{turbofan,turbofandata}. A separate experiment fitted healthy cross-sensor models using 140 training engines from data collection DS02. A threshold selected from the 95th percentile of scores on 30 separate validation engines was fixed before testing 84 engines from collection DS03. Testing contained 16,800 rows, of which 10,707 had active single-sensor drift or step faults.

The turbofan experiment predicts each of fourteen sensors from four operating variables that are not given synthetic faults and the other thirteen sensors. It divides prediction errors by their training standard deviations, selects the sensor with the largest adjusted error after an alarm, and reconstructs that sensor. Models are fitted on the turbofan training data. This experiment evaluates sensor detection, identification, and reconstruction using its own evaluation procedure, rather than the full diesel diagnosis system or the benchmark's official false-alarm protocol.

\begin{table}[htbp]
\centering\caption{Turbofan detection by actual corruption size in units of healthy prediction-error standard deviation. Active drift labels include small changes near fault onset.}
\label{app:turbofan-size}
\begin{tabular}{lrr}
\toprule
Actual corruption size & Active-fault rows & Detected\\
\midrule
Below 0.5 & 1,752 & 4.5\%\\
0.5 to 1 & 1,381 & 4.8\%\\
1 to 2 & 2,239 & 8.8\%\\
2 to 3 & 1,513 & 9.6\%\\
3 to 5 & 1,491 & 75.7\%\\
5 to 8 & 1,550 & 100.0\%\\
8 to 12 & 753 & 100.0\%\\
12 to 20 & 28 & 100.0\%\\
\midrule
All active-fault rows & 10,707 & 36.9\%\\
\bottomrule
\end{tabular}
\end{table}
\FloatBarrier

Alarms occurred on 5.1\% of healthy rows. Among rows with active faults that raised an alarm, sensor identification was correct in 89.4\% of cases. Table~\ref{app:turbofan-size} shows detection rates by the size of the injected change, with smaller changes detected less often.

Reconstruction is evaluated only on rows with active faults that raise an alarm. Attempted replacements reduce summed absolute error by 83.3\%, while 2.5\% increase the selected reading's error. Undetected faults are outside this reconstruction denominator, and corrections following false alarms on healthy rows were not evaluated.

\begin{samepage}
Resampling complete test-engine trajectories 400 times, with models and thresholds fixed, gives intervals of 28.99--44.97\% for detection, 85.38--92.12\% for sensor identification among alarms, 73.34--93.74\% for error removal among attempted replacements, and 0.10--7.69\% for replacements that increase error.

\end{samepage}

\FloatBarrier
\section{Code availability}
\label{app:availability}

Code is available at \href{https://github.com/Tejas7007/NeurIPS-2026-TCCML-Sensor-Diagnosis}{\nolinkurl{github.com/Tejas7007/NeurIPS-2026-TCCML-Sensor-Diagnosis}}.

\end{document}